\documentclass{article}
\usepackage{arxiv}
\usepackage{natbib}
\usepackage{times}

\usepackage{soul}
\usepackage{url}
\usepackage[hidelinks]{hyperref}
\usepackage[utf8]{inputenc}
\usepackage[small]{caption}
\usepackage{graphicx}
\usepackage{amsmath}
\usepackage{booktabs}
\usepackage{xcolor}
\usepackage{hyperref}
\usepackage{url}
\usepackage{graphicx}
\usepackage[linesnumbered, ruled, vlined]{algorithm2e}
\usepackage{algorithmic}

\DeclareTextFontCommand{\celia}{\bfseries \color{blue}}

\title{Towards Creativity Characterization of Generative Models via Group-based Subset Scanning}

\author{Celia Cintas \\
IBM Research Africa\\
Nairobi, Kenya\\
\And
Payel Das\\
IBM Research\\
Yorktown Heights, NY, USA\\
\And 
Brian Quanz\\
IBM Research\\
Yorktown Heights, NY, USA\\
\And 
Girmaw Abebe Tadesse  \\
IBM Research Africa\\
Nairobi, Kenya\\
\And
Skyler Speakman\\
IBM Research Africa\\
Nairobi, Kenya\\
\And
Pin-Yu Chen\\
IBM Research\\
Yorktown Heights, NY, USA\\
}

\begin{document}

\maketitle

\begin{abstract}
Deep generative models, such as Variational Autoencoders (VAEs) and Generative Adversarial Networks (GANs), have been employed widely in computational creativity research. However, such models discourage out-of-distribution generation to avoid spurious sample generation, thereby limiting their creativity. Thus, incorporating research on human creativity into generative deep learning techniques presents an opportunity to make their outputs more compelling and human-like.  As we see the emergence of generative models directed toward creativity research, a need for machine learning-based surrogate metrics to characterize creative output from these models is imperative. 
We propose group-based subset scanning to identify, quantify,  and characterize creative processes by detecting a subset of anomalous node-activations in the hidden layers of the generative models. 
Our experiments on the standard image benchmarks, and their ``creatively generated'' variants, reveal that the proposed subset scores distribution is more useful for detecting novelty in creative processes in the activation space rather than the pixel space. Further, we found that creative samples generate larger subsets of anomalies than normal or non-creative samples across datasets. The node activations highlighted during the creative decoding process are different from those responsible for the normal sample generation. Lastly, we assess if the images from the subsets selected by our method were also found creative by human evaluators, presenting a link between creativity perception in humans and node activations within deep neural nets.

\end{abstract}

\section{Introduction}
\begin{figure*}[t]
    \centering
    \includegraphics[width=0.92\textwidth]{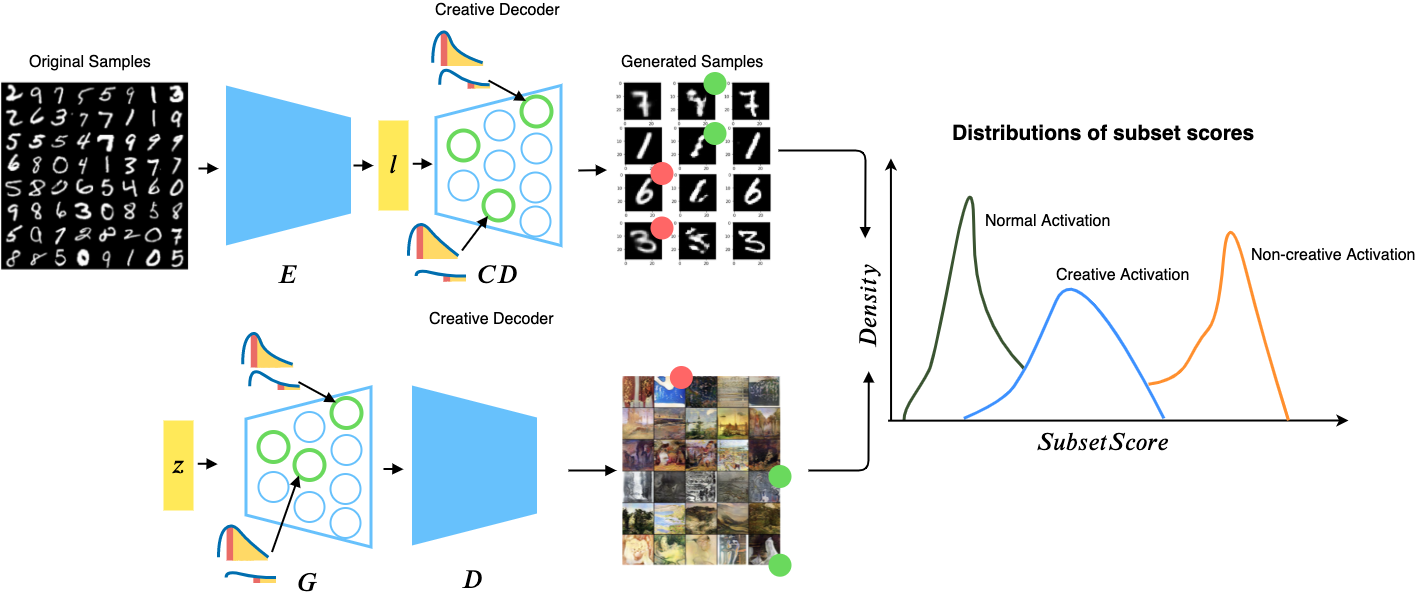}
    \caption{Overview of the proposed approach. First,  we analyze the distribution of the activation space of generative model (e.g. the Creative Decoder $CD$ or a generator $G$ from a GAN). After we extracted the activations from the model for a set of latent vectors $l$, we compute the empirical $p-$values followed by the maximization of non-parametric scan statistics (NPSS). Finally, distributions of subset scores for creative, non-creative processes are estimated, a subset of samples and the corresponding anomalous subset of nodes in the network are identified.}
    \label{fig:overview}
\end{figure*}
Creativity is a process that provides novel and meaningful ideas~\cite{boden2004creative,said2017approaches}. Current deep learning approaches open a new direction enabling the study of creativity from a knowledge acquisition perspective. Novelty generation using powerful deep generative models, such as Variational Autoencoders (VAEs)~\cite{kingma2016improved,rezende2015variational}  and Generative Adversarial Networks (GANs)~\cite{goodfellow2014generative}, have been attempted. 
However, such models discourage out-of-distribution generation to avoid instability and decrease spurious sample generation, limiting their creative generation potential. One of the most controversial issues is the measurement of creativity~\cite{said2017approaches}. Notably, in the machine learning domain, designing creativity evaluation schemes for creative processes and outputs is as essential as developing creative generative methods.
Multiple aspects of creativity, such as expected type (transformational or combinatorial), criteria to assess both the process and the generated output (novelty, value, etc.)~\cite{franceschelli2021creativity} and quantitative metrics need to be better defined to allow the research community to develop and test hypotheses systematically~\cite{cherti2017out}.

This paper proposes a method designed to detect and characterize when a generative deep neural net model produces a creative artifact as per a human evaluator. We employ group-based scanning to determine whether a given batch of generated processes contains creative samples by using an anomalous pattern detection method called group-based subset scanning~\cite{mcfowland-fgss-2013}. 

In short, this work identifies which, of the exponentially-many, subset of samples has higher-than-expected activations at which, of the subset of nodes in a hidden layer of a generative model.   For that purpose, we present experiments with a ``Creative'' variant ~\cite{das2019toward} of a VAE decoder and of an ArtGAN generator ~\cite{artgan2018}.
Furthermore, we provide a quantitative, as well as a qualitative, assessment under the WikiArt dataset and the creative samples guided selection process to evaluate the possibility of using group-based scanning as a proxy for the novelty component in creativity perception.

The main contributions of this paper are: 
\begin{itemize}
    \item We propose an approach to detect and characterize the creative processes and artifacts, within the framework of off-the-shelf generative models, by applying group-based subset scanning methods on node activations from internal layers. 
    \item We present the unique ability to identify patterns of anomalous activations across a group of artifacts.
    \item We evaluate the selected subgroups of artifacts by our proposed method with quantitative surrogate and detection power metrics. Further, we conduct a validation study with a group of human evaluators to qualitatively assess the ``creative'' quality of the selected samples with respect to human perception. 
\end{itemize}

\section{Related Work}
There are multiple surrogate metrics for novelty~\cite{wang2018generative,ding2014experimental,kliger2018novelty} in the literature that commonly used as a proxy for creativity; however, the ultimate test of creativity is done by human inspection.  Human labelling has been used to evaluate deep generative models~\cite{lopez2018human} or as a part of the generative pipeline ~\cite{lake2015human,salimans2016improved}. Although human judgment of creativity has numerous drawbacks,  such as annotation is not feasible for large datasets due to its labor-intensive nature, operator fatigue, and intra/inter-observer variations related to subjectivity, it is still crucial to check how humans perceive and judge generated artifacts~\cite{lamb2018evaluating}. 

Among the surrogate metrics of novelty, for example, In-domain Scores (IS) from a classic novelty detector, such as a One-class SVM or Isolation Forests~\cite{liu2008isolation}, have been a popular choice. Commonly, novel classes are often absent during training, poorly sampled, or not well defined. We can classify each generated image to get an in-domain score (IS) - the signed difference from the normal-classifying hyperplane in the case of One-class SVM or the mean anomaly score of the trees in the forest. The lower this value is, the stronger the outlierness of the image. Similarly, reconstruction distance of any novel image using trained autoencoder models will be larger compared to known samples.

Earlier studies mainly focused on estimating generated samples' novelty without explicitly considering the creativity aspect from a  human perception perspective.~\cite{grace2015data,Grace2019} argue that the novelty component of creativity should be considered relative to observers' expectations rather than based solely on distance in the latent space of a generative model. Further, those novelty measures do not connect with the internal features of the generative model in a quantitative manner, which can provide explanation of the creative generation process. We propose a method to quantify the link between anomalous node activations in inner layers of generative processes and human perception of creativity.
\section{Group-based Subset Scanning Over The Creative Decoder Activation Space}\label{sec:groupscan}

A visual overview of the proposed approach is shown in Figure~\ref{fig:overview}. 
Subset scanning treats the 
creative quantification and characterization problem as a search for the {\em most anomalous} subset of observations in the data, this could be pixels or activations from a given layer of the generator.  The corresponding exponentially large search space is efficiently explored by exploiting mathematical properties of our measure of anomalousness. Consider a set of samples from the latent space  $X =\{X_1 \cdots X_M\}$ and nodes $O = \{O_1 \cdots O_J\}$ within the creative decoder $CD$ or $G$ (See Figure~\ref{fig:overview} for a graphical reference). Where $CD$ and $G$ is a generative neural network capable of producing creative outputs~\cite{das2019toward}. 
Let $X_S \subseteq X$ and $O_S \subseteq O$,  we then define the subsets $S$ under consideration to be $S = X_S\times O_S$. The goal is to find the most anomalous subset:%
\begin{equation}
    S^{*}=\arg \max _{S} F(S)
\end{equation}
where the score function $F(S)$ defines the anomalousness of a subset of samples from the node activations of a given layer from $CD$ or $G$. Group-based subset scanning uses an iterative ascent procedure that alternates between two steps: a step identifying the most anomalous subset of samples for a fixed subset of nodes, or a step that identifies the converse.
There are $2^M$ possible subsets of samples, $X_S$, to consider at these steps.  However, the Linear-time Subset Scanning property (LTSS) \cite{neill-ltss-2012,speakman_penalized} reduces this space to only $M$ possible subsets while still guaranteeing that the highest scoring subset will be identified.  This drastic reduction in the search space is the key feature that enables subset scanning to scale to large networks and sets of samples.  

\subsection{Non-parametric Scan Statistics (NPSS)} Group-based subset scanning uses 
NPSS that has been used in other pattern detection methods \cite{mcfowland-fgss-2013,feng-npss_graph-2014,cintas2020detecting}. Given that NPSS makes minimal assumptions on the underlying distribution of node activations, our approach has the ability to scan across different type of layers and activation functions. 
There are three steps to use the non-parametric scan statistics on the model's activations. The first is to form a distribution of  ``expected''  activations at each node ($H_0$). We generate this distribution by letting the generative process create samples that are known to be from the training data (sometimes referred to as ``background'' samples) and record the activations at each node. The second step involves scoring a group of samples in a test set that may contain creative or normal artifacts. We 
record the activations induced by the group of test samples and compare them to the baseline activations created in the first step. This comparison results in a $p$-value for each sample in the test set at each node.  
Lastly, we quantify the anomalousness of the resulting $p$-values by finding $X_S$ and $O_S$ that maximize the NPSS, which estimates  how much an observed distribution of $p$-values deviates from the uniform distribution. 

Let $A^{H_0}_{zj}$ be the matrix of activations from $l$ latent vectors of training samples at each of $J$ nodes in a creative decoder layer.  Let $A_{ij}$ be the matrix of activations induced by $M$ latent vectors in the test set, that may or may not be novel.  
Group-based subset scanning computes an empirical $p$-value for each $A_{ij}$, as a measurement for how anomalous the activation value of a potentially novel sample $X_i$ is at node $O_j$. 
This $p$-value $p_{ij}$ estimates the proportion of activations from the $Z$ background samples, $A^{H_0}_{zj}$, that are larger or equal to the activation from an evaluation sample 
at node $O_j$.
\begin{equation}
    p_{ij} = \frac{1+\sum_{z=1}^{|Z|} I(A^{H_0}_{zj} \geq A_{ij} )}{|Z|+1}
\end{equation}
Where $I(\cdot)$ is the indicator function. A shift is added to the numerator and denominator so that a test activation that is larger than \emph{all} activations from the background at that node is given a non-zero $p$-value.  Any test activation smaller than or tied with the smallest background acivation at that node is given a $p$-value of 1.0. 

Group-based subset scanning processes the matrix of  $p$-values ($P$) from test samples with a NPSS to identify a submatrix $S =X_S \times O_S$ that maximizes  $F(S)$, as this is the subset with the most statistical evidence for having been affected by an anomalous pattern. 
The general form of the NPSS score function is 
\begin{equation}
F(S)=\max_{\alpha}F_{\alpha}(S)=\max_{\alpha}\phi(\alpha,N_{\alpha}(S),N(S))
\end{equation}
where $N(S)$ is the number of empirical $p$-values contained in subset $S$ and $N_{\alpha}(S)$ is the number of $p$-values less than (significance  level) $\alpha$ contained in subset $S$. 
It has been shown that for a subset $S$ consisting of $N(S)$ empirical $p$-values, $E\left[N_{\alpha}(S)\right] = N(S)\alpha$ ~\cite{mcfowland-fgss-2013}.
Group-based subset scanning attempts to 
find the subset $S$ that shows the most evidence of an observed significance higher than an expected significance, 
$ N_{\alpha}(S) > N(S)\alpha $, for some significance level $\alpha$.

In this work, we use the Berk-Jones (BJ) test statistic as our scan statistic. BJ test statistic is defined as:
\begin{equation}
    \phi_{BJ}(\alpha,N_\alpha,N) = N*{KL} \left(\frac{N_\alpha}{N},\alpha\right)
\end{equation}
where $KL$ refers to the Kullback-Liebler divergence, $KL(x,y) = x \log \frac{x}{y} + (1-x) \log \frac{1-x}{1-y}$, between the observed and expected proportions of significant $p$-values. We can interpret BJ as the log-likelihood ratio for testing whether the $p$-values are uniformly distributed on $[0,1]$.
\begin{figure*}[h!]
    \centering
 \includegraphics[width=1.\textwidth]{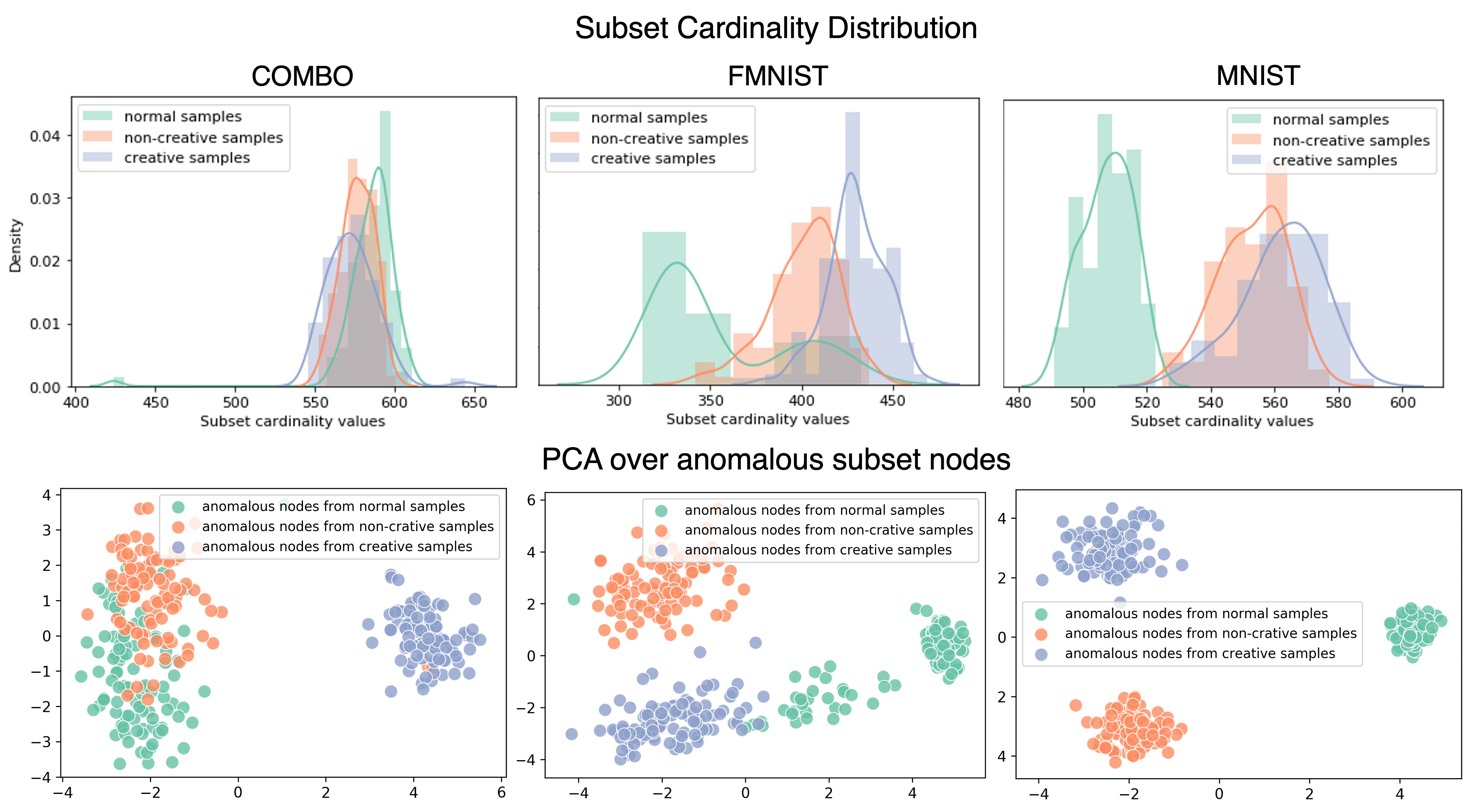}
    \caption{Characterization of activations across the three datasets. \textbf{(top)} Subset cardinality distributions for anomalous subsets for different type of generated samples. \textbf{(bottom)} PCA over anomalous subset nodes for the creative decoder activations under generation of normal, non-creative and creative samples. }
    \label{fig:cardinalitydist}
\end{figure*}

\section{Experimental Setup}
We hypothesize that creative content leaves a subtle but systematic trace in the activation space that can be identified by looking across multiple creative samples. Further, we assume that not all generative models will have the same throughput of creative samples in a batch. Thus, we need to evaluate our method under different proportions to see if even models that generate a small percentage of creative samples can be detected by our method. 
We test this hypothesis through group-based subset scanning over the activation space that encodes \emph{groups of artifacts} that may appear anomalous when analyzed together.
Given that the proposed approach is model agnostic, we test it under two generative models, a creative VAE Decoder and a Creative Generator variant of ArtGAN architectures~\cite{das2019toward,artgan2018}. We scan both the pixel and activation space. We also validate the proposed approach across multiple datasets, including images from MNIST, Fashion-MNIST (FMNIST) \cite{xiao2017fashion}, Combo~\cite{das2019toward}, which is a combination of the previous two datasets, and WikiArt (with ArtGAN)~\cite{artgan2018}.  

In our experiments we quantify detection \emph{power}, that is the method's ability to distinguish between test sets that contain some proportion of creative samples and test sets containing only normal content, using Area Under the Receiver Operating Characteristic Curve (AUROC) (See Figure~\ref{fig:eval_batch}B). Furthermore, we compute multiple existing surrogate metrics of novelty (e.g. in-domain score and reconstruction error, see Related Work) in the groups of artifacts found by our method to compare the novelty score of the found samples under group-based subset scanning. Further, for the WikiArt dataset, we conducted a questionnaire across multiple human evaluators to understand what makes a sample creative under this dataset and see if there is a correspondence between the anomalous subset selected by our method and the creativity perception of the evaluator. 

\subsection{Datasets and Creative Labelling}
For human evaluation under FMNIST, MNIST, and Combo datasets, we use the labeled artifacts from ~\cite{das2019toward} to test the performance of our proposed method. Nine evaluators did the human annotation over a pool of 500 samples per dataset (we used agreement amongst $>3$ annotators as consensus), generated from either the creative generative process or regular decoding. 

To evaluate the creative perception and our approach for a guided ``creative'' sampling process, we follow~\cite{amabile1982social}, this technique measures creativity using an set of evaluators, who assess creative works individually and in isolation. The evaluators' feedback is then collected and aggregated to establish an overall rating or measure. In our preliminary study, ten voluntary evaluators completed a questionnaire evaluating three sets of eleven images under the WikiArt dataset with ArtGAN, an example of one set can be found in Figure~\ref{fig:eval_batch}A. We show the baseline generated by standard ArtGAN and the sampled subset of artifacts selected by our method in the online form. These samples can belong to the ArtGAN output or the creative variant. We ask if the sample is creative compared to the initial image (See Figure~\ref{fig:eval_batch}C). In this case, given the dataset complexity we used agreement amongst $>5$ annotators as consensus. Further, we ask each evaluator to provide a short description of what makes a sample creative and non-creative, to better understand  how the evaluator assess the creative aspect of a sample.

\subsection{Subset Scanning and Base Models Setup}
We use a VAE~\cite{kingma2016improved} and ArtGAN~\cite{artgan2018} as the base generative models, and 
the ``low-active'' method proposed in ~\cite{das2019toward} as the creative process for our experiments. The ``low-active'' method identifies neurons that typically have low activation across all the training data and turns some number of them on at decoding time.

We run individual and group-based scanning on node activations extracted from the Creative VAE Decoder and the Creative ArtGAN  Generator. We tested group-based scanning across several proportions of creative content in a group, ranging from $10\%$ to $50\%$. We used $Z=250$ latent vectors to obtain the background activation distribution ($A^{H_0}$) for experiments with both datasets. For evaluation, each test set had samples drawn from a set of $100$ normal samples from the regular decoder (separate from Z) and $100$ samples labeled as creative and $100$ non-creative samples (not novel or creative label).

\begin{table}[h]
\centering

\begin{tabular}{lcccc}
\toprule
\multicolumn{1}{c}{Space} &
\multicolumn{1}{c}{Dataset} & \multicolumn{3}{c}{Subset Scanning}\\
      \midrule
      & & \multicolumn{1}{c}{50\%} &
       \multicolumn{1}{c}{10\%} &
      \multicolumn{1}{c}{Indv.} \\
 \midrule
Pixel Space & MNIST & $0.971$    & $0.791$             & $0.255$ \\
Activation Space &  MNIST & $\mathbf{0.991}$    & $\mathbf{0.972}$             & $0.531$  \\ 
\midrule
Pixel Space & FMNIST & $0.952$    & $0.743$            & $0.381$  \\ 
Activation Space &  FMNIST & $\mathbf{0.990}$  & $\mathbf{0.962}$           & $0.596$ \\ 
\midrule
Pixel Space & Combo & $0.921$ & $0.713$ & $0.490$ \\
Activation Space & Combo & $\mathbf{0.998}$ & $\mathbf{0.949}$ & $0.626$ \\
\bottomrule
\end{tabular}
 \caption{Detection Power (AUC) for group-based and individual subset scanning over pixel and activation space for the Creative Decoder.} 
\label{table:ssovercreativedecoder}
\end{table}

\section{Results} 
\begin{figure*}[t!]
    \centering
    \includegraphics[width=0.9\textwidth]{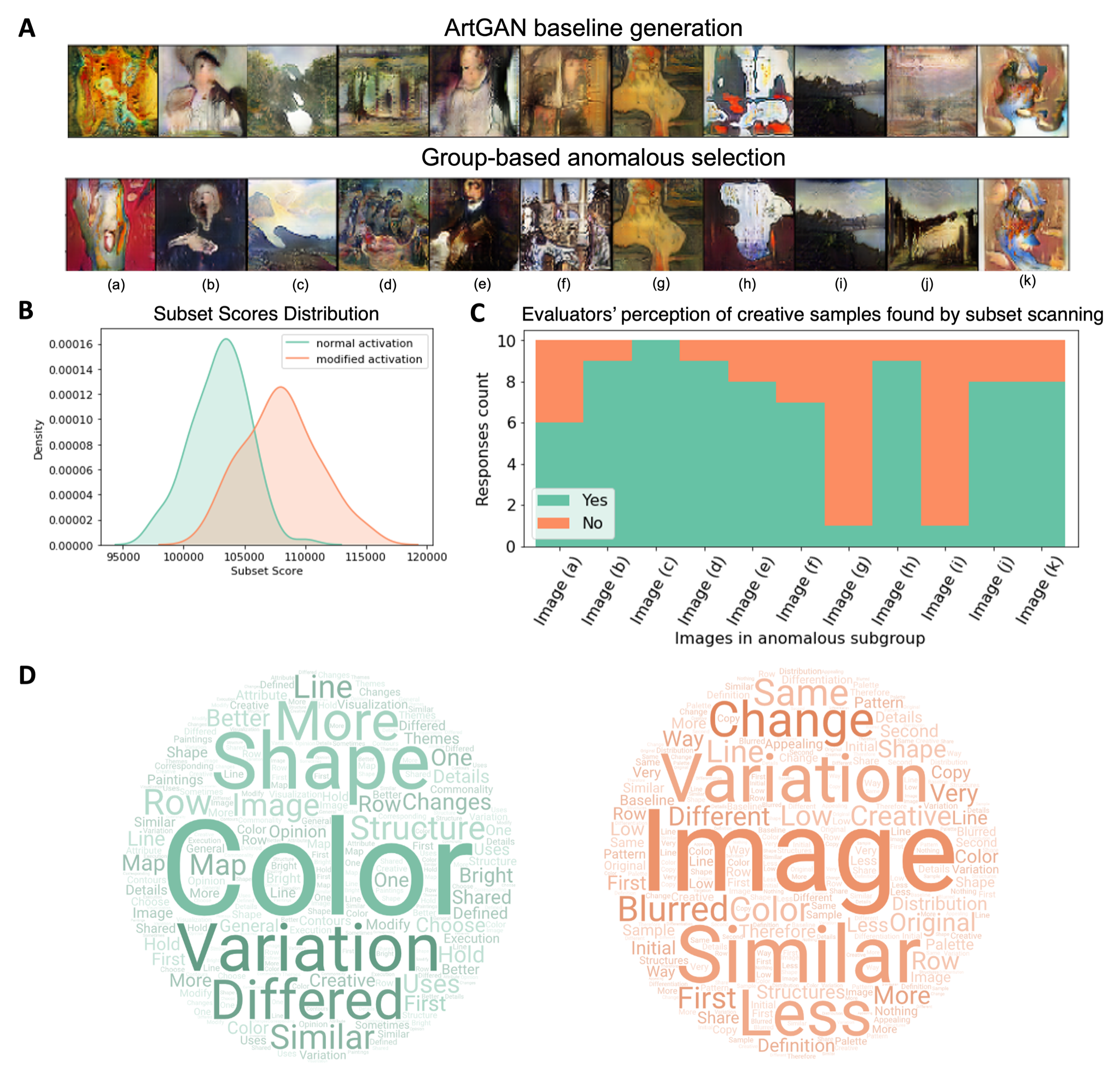}
    \caption{{\bf A)} Batch example used in the online questionnaire for WikiArt dataset. We show the baseline generated by ArtGAN and the guided selected subset of samples that can belong to the ArtGAN output or the creative modification. {\bf B)} Subset score distributions for ``normal'' and creative modified activations. {\bf C)} Binary responses from evaluators for batch displayed in block A, Yes: Perceived as creative, No: Not perceived as creative sample. {\bf D)} Word Cloud visualization from responses to {\em What made the sample creative in your opinion?} (left) and {\em What made the sample non-creative in your opinion? } (right).}
    \label{fig:eval_batch}
\end{figure*}
In Table~\ref{table:ssovercreativedecoder} we present results showing the creative detection capabilities of both activation and pixel spaces. We see that the characterization improves when detecting the creative samples in the activation space, than when we scan over the pixel space. One can also observe that, even when the creative model generates a low throughput (10\%) of creative samples, the proposed method is capable of detecting these samples. 

Additionally,  as shown in Figure~\ref{fig:cardinalitydist}, for the creative-labeled datasets we observe a higher cardinality of anomalous nodes during creative generation, when  compared to normal and non-creative in FMNIST and MNIST. This observation is consistent with the basic principle of the creative decoding process~\cite{das2019toward}, which induces a neuro-insprired atypical neuronal activation in the neural net generator to promote creativity. To further inspect the activations, we visualize the principal component projections of the anomalous subset of nodes for  different sets of  samples. As we can see, the  activations for different types of samples are distinctive. Notably, for FMNIST and Combo we start noticing some overlap for normal and creative samples. Based on this observation, we hypothesize that as more complex datasets are subject to creative decoding, we will see appearance of more overlapping nodes.

\begin{figure*}[h!]
    \centering
    \includegraphics[width=0.95\textwidth]{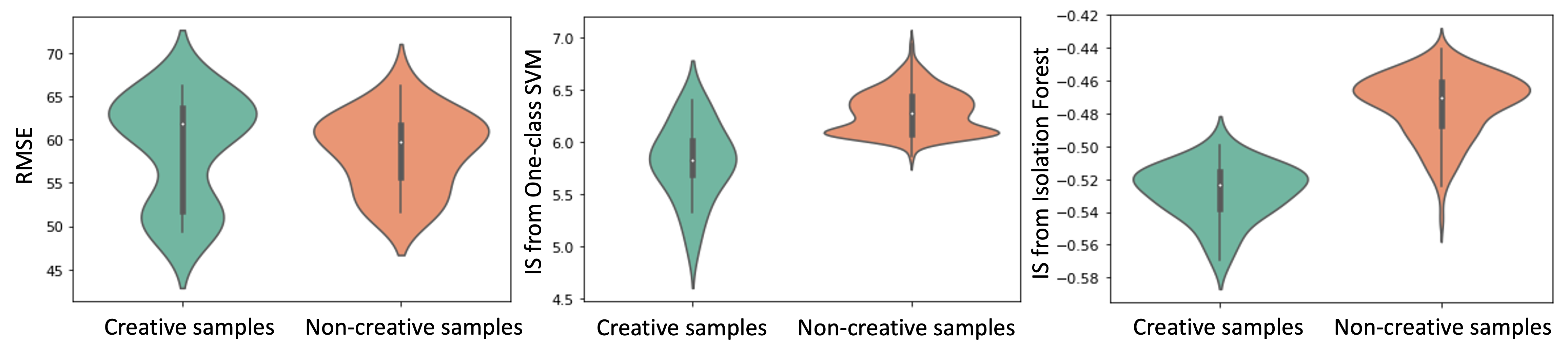}
    \caption{Violin plots for novelty scores across a batch of artifacts sampled with group-based subset scanning from WikiArt dataset.}
    \label{fig:surrogate_violins}
\end{figure*}
\subsection{Creativity Art Perception in Anomalous Subgroups}

Due to the lack of labeled samples in term of creativity in the generated WikiArt, we performed an unsupervised experiment (See Table~\ref{table:ssoverartgan}) that employed extraction of the anomalous subset of images and activations in the generation process. Given the missing labels, we used the generated subset score distribution for anomalous activations (See Figure~\ref{fig:eval_batch}B) as a guide for generating ``creative'' samples (See Figure~\ref{fig:eval_batch}A). This set of outputs was evaluated with several surrogate metrics of novelty (See Table~\ref{table:surrogate} and Figure~\ref{fig:surrogate_violins}). We observe that IS scores have higher values for non-creative samples, and lower scores for the samples selected by our guided ``creative'' sampling method.

Additionally,  the anomaly-guided generations were evaluated with respect to creativity by a group of  human observers  (Figure~\ref{fig:eval_batch}C). Across the three example batches, on average 78\% of the images  were consistently found as creative by evaluators' perception. Apart from requesting the evaluator to tag a sample as creative or not, we requested a brief description of what makes the sample creative and non-creative. The word clouds containing all responses for each question from evaluators can be seen in Figure~\ref{fig:eval_batch}D. Non-creative modification were usually described by evaluators as, \textit{similar to the base image}, \textit{blurred artifacts}, or \textit{not holding any structure with the initial image}, and \textit{less variation of shapes and colors}. While creative samples were described with changing one property (\textit{different color palette} or \textit{variation on the structure}).
\begin{table}[tbph]
\centering
\begin{tabular}{lccc}
\toprule
\multicolumn{1}{c}{Space} & \multicolumn{1}{c}{\# nodes}& \multicolumn{2}{c}{Subset Scanning}\\
      \midrule
      & & \multicolumn{1}{c}{50\%} &  \multicolumn{1}{c}{Indv.} \\
 \midrule
Pixel Space  & 16384 &$0.762$    & $0.504$ \\
Layer 1  & $8192$ & $\mathbf{0.905}$    & $0.571$ \\ 
Layer 2   & $32768$ & $\mathbf{0.913}$    & $0.528$ \\ 
\bottomrule
\end{tabular}
 \caption{Detection Power (AUC) for group-based and individual subset scanning over pixel and activation space for Creative ArtGAN Generator.} 
\label{table:ssoverartgan}
\end{table}

\begin{table}[]
\begin{center}

\begin{tabular}{@{}llll@{}}
\toprule
Subset samples & RMSE       & IS-One-SVM & IS-IF \\ \midrule
Creative       & $59.0\pm 5.9$ & $5.81\pm 0.35$        &  $-0.52\pm0.01$    \\
Non-creative   & $58.1\pm 4.2$ & $6.59\pm 0.20$         &  $-0.47\pm0.01$     \\ \bottomrule
\end{tabular}
    
\end{center}
\caption{Comparison between selected subset of WikiArt samples as creative by our method and novelty metrics: Reconstruction distance (RMSE) from original to creative  generated, in-domain score (IS) obtained using one-class SVM classifier and using Isolation Forest (IF).}
\label{table:surrogate}
\end{table}



\section{Conclusion and Future Work}
Our proposed method for creativity detection in machine-generated images works by analyzing the activation space for off-the-shelf generative models (such as Creative VAE Decoder and Creative ArtGAN Generator). We provide both the subset of the generated artifacts identified as creative and the corresponding nodes in the network's activations that identified those samples as creative. We evaluate the proposed approach across  computer vision datasets and different generative models, to understand how we can better capture the human perception of creativity under more complex domains, such as the WikiArt dataset. We assess the anomaly-guided generations  with existing surrogate metrics for novelty, as well as  with a human evaluation study. Results consistently show the ability of nodes with anomalous activations within a deep generative model for identifying the novelty aspect of creative samples.

Currently, the study faces limitations given the binary creative definition for the artifacts. Future research will consider questions to evaluators to provide a disentanglement of creativity properties such as novelty, surprise, and value. Hence,  providing a more refined granularity will help better understanding of the connections between human perception of creativity and the activation space.
Further, we plan to  leverage the proposed creativity quantification approach as a control for more efficient and trusted generation of artifacts that are consistent with human perception of the novelty component of creativity.

\bibliographystyle{named}
\bibliography{ijcai22}

\end{document}


\maketitle


\section{Group-based Subset Scanning Algorithm}
\textbf{Efficient maximization of NPSS} Group-based subset scanning identifies the anomalous subset of $p$-values through iterative ascent of two optimization steps. 
Within each step, the number of subsets to consider is reduced from $O(2^E)$ to $O(E)$ where $E$ is the number of elements currently being optimized, either latent vectors or nodes. 
This efficient optimization is a direct application of the LTSS property \cite{neill-ltss-2012,speakman_penalized}.  Each element $e$ is sorted by its priority, which is its proportion of $p$-values less than an $\alpha$ threshold.  Once sorted, the LTSS property states that the highest-scoring subset will consist of the top-$k$ elements for some $k$ between 1 and $|E|$. Any subset not consisting of the top-$k$ priority elements is sub-optimal and therefore does not need to be evaluated.

Group-based subset scanning identifies the anomalous subset of $p$-values through iterative ascent of two optimization steps, see Algorithm~\ref{alg:singlerestart}. Within each step, the number of subsets to consider is reduced from $O(2^E)$ to $O(E)$ where $E$ is the number of elements currently being optimized, either images or nodes, see Algorithm~\ref{alg:optimize_rows}. 
\begin{algorithm}[htp]
\caption{Single Restart over $M$ test images and $J$ nodes }\label{alg:singlerestart}
\SetKwFunction{OptimizeRows}{OptimizeRows}
\SetKwFunction{Random}{Random}
\SetKwInOut{Input}{input}
\SetKwInOut{Output}{output}

\Input{$(M \times J)$ $p$-values}
\Output{score, $X_s$, $O_s$}
score $\leftarrow -1$ \;
$X_s \leftarrow$ \Random{$M$} \;
$O_s \leftarrow$  \Random{$J$} \;
\While{score is increasing }{
$(M \times J') = (M \times J) | O_s$ \; 
score, $X_s \leftarrow$  \OptimizeRows{($M \times J'$)} \; 
$(M' \times J) = (M \times J) | X_s$ \; 
score, $O_s \leftarrow$  \OptimizeRows{$(J \times M')$}\; 
} 
\Return{score, $X_s$, $O_s$}
\end{algorithm}
\begin{algorithm}[h]
\caption{Optimize over rows using LTSS. It maintains maxscore and $arg\_max\_subset$ over $\|E\|*\|T\|$ subsets.} \label{alg:optimize_rows}
\SetKwFunction{SortByPropCT}{SortByPropCT}
\SetKwFunction{LinearSpace}{LinearSpace}
\SetKwInOut{Input}{input}
\SetKwInOut{Output}{output}
\Input{$p$-values from all rows E and relevant cols C}
\Output{maxscore, $arg\_max\_subset$}
maxscore $\leftarrow -1$\; $arg\_max\_subset \leftarrow \emptyset$ \;
\For{$t$  in T = \LinearSpace{0,1}}{ 
$sorted\_priority \gets$ \SortByPropCT{E, t}  \Comment{/* Sort elements in $E$ by proportion of $p$-values across $C < t$}.   */ 

 $Score(sorted\_priority, t)$  \Comment{/* Score $|E|$ subsets of $sorted\_priority$ by iteratively including elements one at a time. */} \;
}
\Return maxscore, $arg\_max\_subset$
\end{algorithm}
\section{Further Experiments on Subset scores}
\begin{figure*}[h!]
    \centering
    \includegraphics[width=0.5\textwidth]{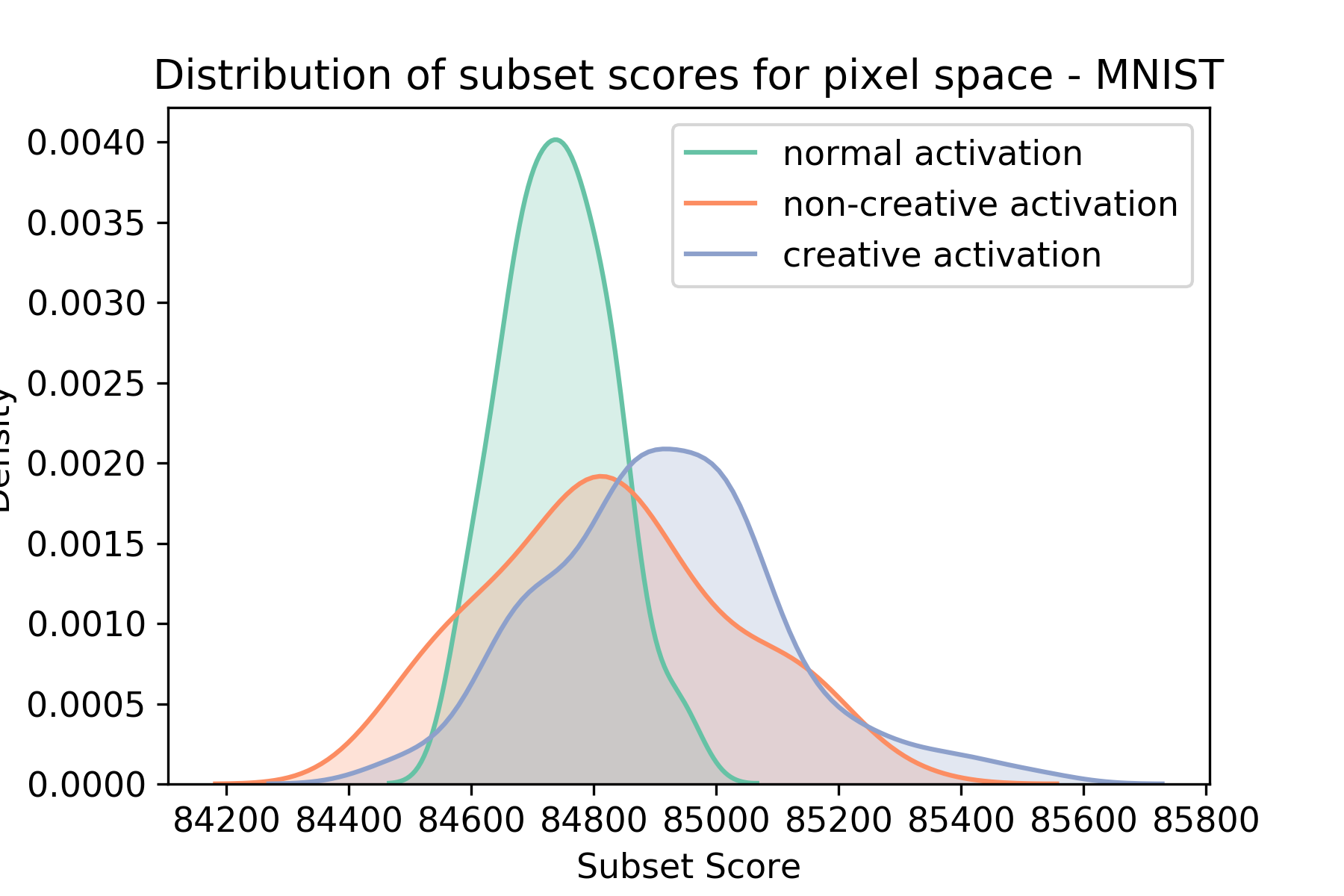}\includegraphics[width=0.5\textwidth]{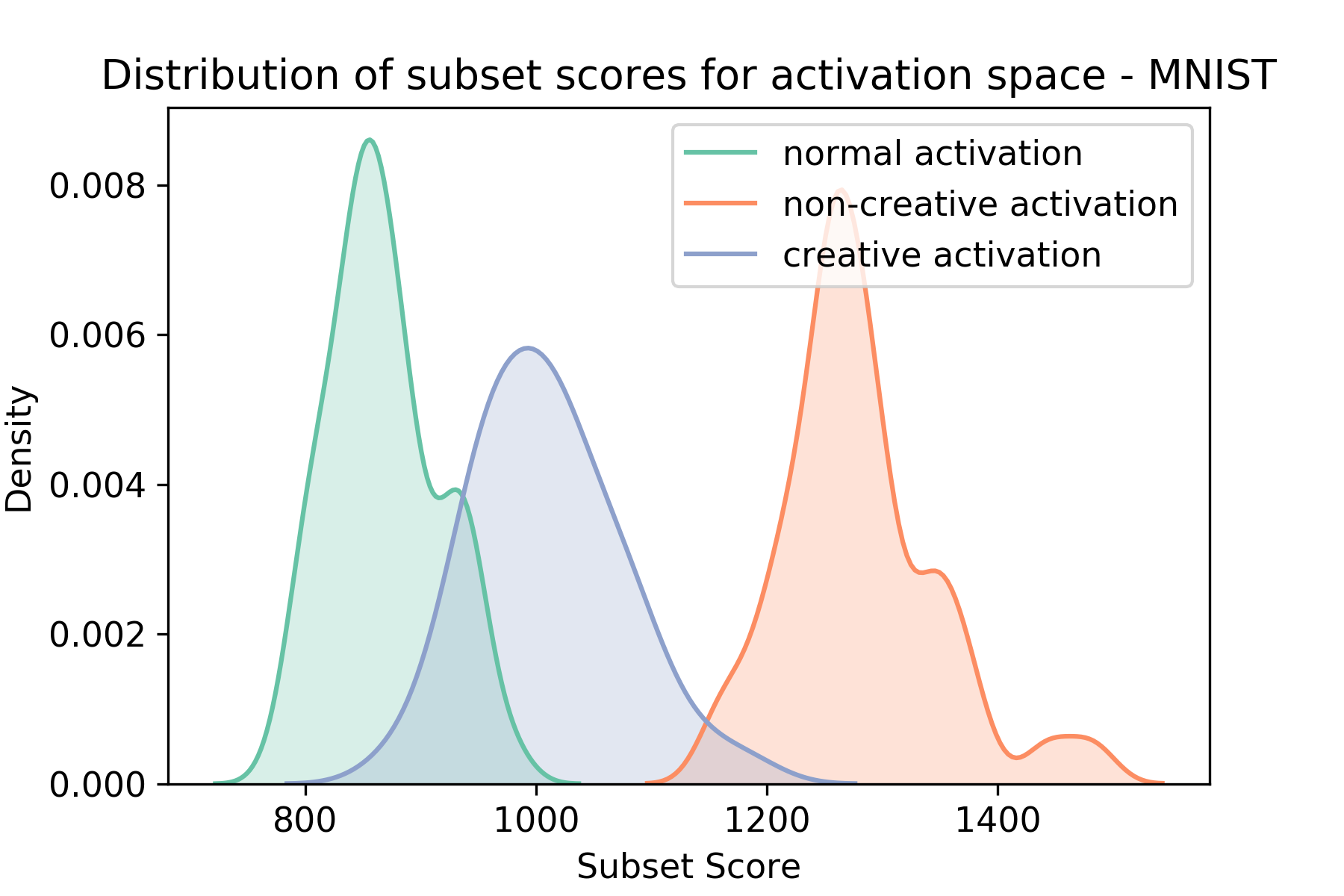}
    \caption{Subset Scores Distributions for normal, non-creative and creative labeled samples under the pixel space (left) and over creative samples in the activation space of the $CD$ (right).}
    \label{fig:subsetscores}
\end{figure*}

The distributions of subset scanning scores are shown in Figure~\ref{fig:subsetscores}, in green for normal samples (expected distribution), in orange for non-creative samples and blue for creative samples. Higher AUCs are expected when distributions are separated from each other and lower AUCs when they overlap. The computed AUC for the subset score distributions can be found in Table 1 (from the main paper).

\section{ArtGAN extended results}
In Figure~\ref{fig:artgan} we can observe samples generated under creative modifications and the corresponding subset scores across each layer.
\begin{figure*}[h!]
    \centering
    \includegraphics[width=\textwidth]{images/artgan.png}
    \caption{{\bf A} Examples of generated samples with creative modifications in Layer 1, Layer 2 and the original generated samples by ArtGAN. {\bf B} Subset Score Distributions and reported AUC for individual and group-based scanning across both layers.}
    \label{fig:artgan}
\end{figure*}

\section{Creativity Questionnaire}
\begin{figure*}[h!]
    \centering
    \includegraphics[width=\textwidth]{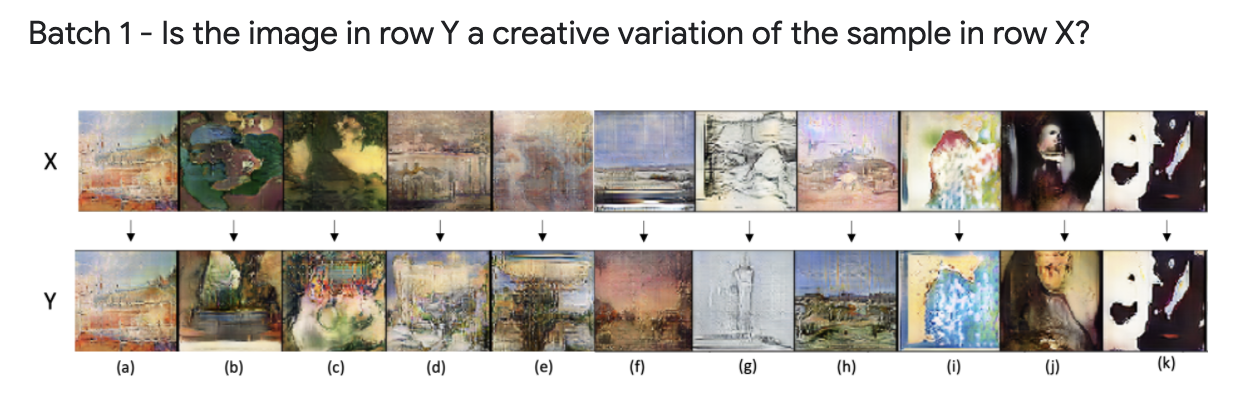}
    \includegraphics[width=\textwidth]{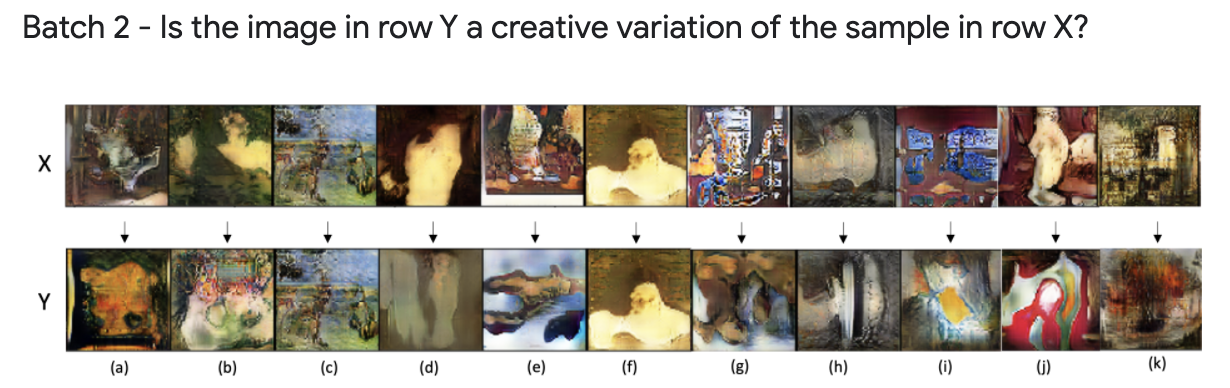}
    \includegraphics[width=\textwidth]{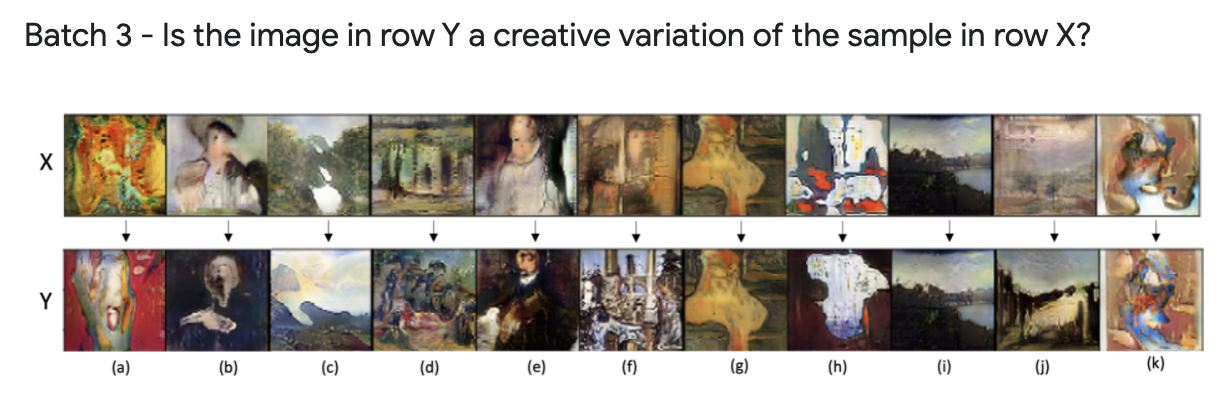}
    \caption{Batches of samples used in the online questionnaire. Apart from a binary response on each sample we added two questions to understand which properties made a sample creative.}
,    \label{fig:questionnaire}
\end{figure*}

For the human evaluation we use three randomly selected batches of output from our proposed method. The images can be seen in Figure~\ref{fig:questionnaire}.

